\documentclass[nohyperref]{article}

\usepackage{microtype}
\usepackage{subcaption}
\usepackage{graphicx}
\usepackage[table,xcdraw]{xcolor}
\usepackage{booktabs} 

\usepackage{hyperref}

\usepackage[accepted]{icml2022}

\usepackage{comment}

\usepackage{amsmath}
\usepackage{amssymb}
\usepackage{mathtools}
\usepackage{amsthm}
\usepackage{fdsymbol}
\usepackage{nicefrac}
\usepackage{bbm}

\usepackage[capitalize,noabbrev]{cleveref}

\theoremstyle{plain}

\theoremstyle{definition}

\theoremstyle{remark}

\usepackage[textsize=tiny]{todonotes}

\newcommand{\oracle}{\ensuremath\mathcal{O}}
\newcommand{\budget}{\ensuremath s_{\max}}

\icmltitlerunning{Quantity vs quality}

\begin{document}

\twocolumn[
\icmltitle{Quantity vs Quality: Investigating the Trade-Off between \\Sample Size and Label Reliability}



\icmlsetsymbol{equal}{*}

\begin{icmlauthorlist}
\icmlauthor{Timo Bertram}{at}
\icmlauthor{Johannes Fürnkranz}{at}
\icmlauthor{Martin Müller}{ca}
\end{icmlauthorlist}

\icmlaffiliation{at}{Institute for Application-oriented Knowledge Processing, Johannes Kepler University Linz, Austria}
\icmlaffiliation{ca}{Department of Computer Science, University of Alberta, Canada}

\icmlcorrespondingauthor{Timo Bertram}{tbertram@faw.jku.at}

\icmlkeywords{Machine Learning, Online Learning, Labeling, MNIST, Oracle}

\vskip 0.3in
]



\printAffiliationsAndNotice{} 

\begin{abstract}
In this paper, we study learning in probabilistic domains where the learner may receive incorrect labels but can improve the reliability of labels by repeatedly sampling them. In such a setting, one faces the problem of whether the fixed budget for obtaining training examples should rather be used for obtaining all different examples or for improving the label quality of a smaller number of examples by re-sampling their labels. 
We motivate this problem in an application to compare the strength of poker hands where the training signal depends on the hidden community cards, and then study it in depth in an artificial setting where we insert controlled noise levels into the MNIST database. 
Our results show that with increasing levels of noise, resampling previous examples becomes increasingly more important than obtaining new examples, as classifier performance deteriorates when the number of incorrect labels is too high. In addition, we propose two different validation strategies; switching from lower to higher validations over the course of training and using chi-square statistics to approximate the confidence in obtained labels.
\end{abstract}

\section{Introduction}
\label{sec:Introduction}

Sample efficiency is one of the main concerns when training neural networks. Having a better sample efficiency leads to faster training and, when using online learning, requires less computational effort to generate new samples. A different but related aspect of deep learning is the reliability of labels. Noisy or incorrect labels hinder training and often lead to slower convergence and worse performance \cite{song2020learning}. Therefore, label reliability and the efficiency of training are inherently related \cite{reed2014training}. 

While the reliability of labels can not be influenced in some contexts, we consider a problem setting where we have a fixed budget of samples that can be used for training a classifier. At each step after receiving the label of a sample, we can decide whether we want to sample a new example or resample the label of the previously seen example in order to validate the correctness of the label.
This can, e.g., be used in noisy domains, where the observed label has a certain error probability, or in probabilistic problems, where each example is associated with a probability distribution over the labels.
For simplicity, we will in the following typically assume that we have a single correct label which may be distorted by noise in the data, but the same setting is also applicable to probabilistic domains.
Practical applications of this include a crowd-sourcing campaign where letting multiple participants label a sample will increase the likelihood of the label being correct, 
or probabilistic domains, where the label depends on a probabilistic event. We use one such case, poker, as a motivating example, where the task is to predict the winner of two hands, but the outcome depends on hidden community cards.

In this work, we do not provide the classifier any information about the noisiness of the data. In addition, we intentionally do not take any noise-reducing measures apart from the validation process. For both series of experiments, we use the neural network architecture that performed well on the non-noisy variant for the experiments with noise. This noise-agnostic approach was chosen such that noise-handling in the classifier and training algorithm is decoupled from the domain we are investigating here; handling noise in the data generation step.

After a brief discussion of related work (Section~\ref{sec:related}), we motivate the research problem with an experiment in the domain of poker, where the question of whether to re-sample known examples to increase label reliability or to sample new examples for increasing the size of the training set naturally occurs (Section~\ref{section:Poker}). We then formalize the problem and our research questions in Section~\ref{section: Problem definition}, and empirically investigate it in a set of studies on the MNIST data where controlled levels of noise were inserted (Section~\ref{section:MNIST}). The conclusions that we draw from these experiments are summarized in Section~\ref{sec:conclusions}.

\section{Related work}
\label{sec:related}

Learning in noisy and probabilistic domains has been studied extensively in the machine learning literature. However, such works mostly cover other solutions than this work aims to, partly due to using different assumptions about receiving training samples. One common approach when handling noisy labels is to omit the questionable labels in training and later use a classifier trained on non-omitted labels as ground truth to identify errors. Through cross-validation, this then yields adjusted labels that can be used for the final training of the network \cite{chen2019understanding}.
Other works use specific layers to handle the noisy data \cite{sukhbaatar2014learning}, which however also requires knowledge about the noise distribution which has to be learned in a previous step. Additionally, previous research aims to change loss functions to account for noise in the data \cite{sukhbaatar2014learning,zhang2018generalized}.

Different related areas of research concern cost-effective and active learning \cite{ActiveLearning,costEffective}, where the algorithm can query an oracle to label specific samples. There, one of the main goals is to train classifiers as efficiently as possible with respect to required samples. Active learning also provides an interesting parallel to this work as both use an oracle to label samples. The main difference between active learning and what is studied here lies in the fact that in this domain, the oracle is noisy and multiple independent queries to the oracle can be used to reduce noise. In active learning, the algorithm aims to identify which samples provide the highest benefit and queries the oracle for the labels of those. While the combination of both settings is not studied here, it provides an interesting ground for further research. Similar to active learning, it may be especially beneficial to validate more important samples further, while less interesting samples may require fewer validations to efficiently progress the classifying performance.

Our work is also related to explicit noise filtering algorithms, such as the ensemble filters introduced in \cite{NoiseFiltering}. Like these algorithms, we can use multiple queries to correct noisy labels, but we assume that we can directly query the labeling oracle instead of using an ensemble of diverse noise-tolerant learners, and we explicitly aim at studying the trade-off between obtaining new examples and re-sampling known examples. 

\section{Motivating Example: Poker}
\label{section:Poker}

Our work is motivated by experiments that we did in
learning to evaluate
hand strength in Texas Hold 'em Poker. The task here is to predict which of two 2-card hands is more likely to win the game after all five community cards are revealed (the 'river'), given the first three community cards (the 'flop'). In this domain, the label quality vs.\ quantity problem occurs naturally: Should we obtain multiple samples of the missing two community cards in order to get a more reliable estimate of which of the two hands is stronger in the context of the current flop, or should we instead increase our training set with more samples, i.e., different combinations of hand and flop cards?

\subsection{Experimental Setup}
We investigate this domain by training neural networks. Thus, a training sample consists of three inputs ($p1$, $p2$ and $flop$) while the output signals whether $p1$ or $p2$ is the better hand in the context of the given flop. To account for such a structure, we train a Siamese network \cite{koch2015siamese,chicco2021siamese}. While the complete Siamese structure that was used can not be explained fully here, Siamese networks can not only be used in one-shot image learning but also to model which of several inputs is preferential to another \cite{tesauro1989connectionist} in a given context \cite{bertram2021comparison}. Here, the two hands and the flops are used as inputs for the Siamese network and its output models which of the two hands is assumed to be better given the flop.

The test accuracy of the network is computed by taking the average over all possible rivers that extend the current flop, i.e., if $P(p1 \textrm{\ wins} | p1, p2, flop) > P(p2 \textrm{\ wins} | p1, p2, flop)$,  then the network should predict that $p1$ is the better hand. For training, the network receives $p1$, $p2$ and the $flop$ as an input. In this situation, the algorithm has the option to choose how often it wants to validate the training signal, i.e. which hand is assumed to be better. One validation consists of drawing two random and legal community cards to form the river and see which of the two hands wins the game. By doing this multiple times, a training signal is constructed which, is less noisy for higher numbers of validations, i.e. as more different rivers are seen, which increases the chance of the assumed better hand to be the better hand in general.

To emphasize this, consider the following example. 
Hand $p1 = (Q\varheartsuit, J\spadesuit)$, hand $p2 = (7\spadesuit, 7\vardiamondsuit)$, and $flop = (2\spadesuit,9\spadesuit,10\spadesuit)$. The algorithm then decides how often it wants to draw two remaining cards to approximate the better hand, e.g. for one draw the river may result in $(2\spadesuit,9\spadesuit,10\spadesuit, Q\vardiamondsuit,K\vardiamondsuit)$, resulting in the approximation that $P(p1 \textrm{\ wins}|p1, p2, flop) = 1$ and $P(p2 \textrm{\ wins}|p1, p2, flop) = 0$, with the network using the information that hand $p1$ is better than hand $p2$ given the flop as the training signal. For two additional validations, those rivers may result in $(2\spadesuit,9\spadesuit,10\spadesuit, A\vardiamondsuit,3\vardiamondsuit)$ and $(2\spadesuit,9\spadesuit,10\spadesuit, 10\vardiamondsuit,10\varheartsuit)$, approximating the winning chances for p1 and p2 as 0.33 and 0.67 respectively. In the limit of validations, this approximation will result in the true win-probabilities (here $P(p1 \textrm{\ wins}|p1, p2, flop) = 0.6697$ and $P(p2 \textrm{\ wins}|p1, p2, flop) = 0.3303)$.

\subsection{Result}

\begin{figure}[t]
    \centering
    \includegraphics[width=\linewidth]{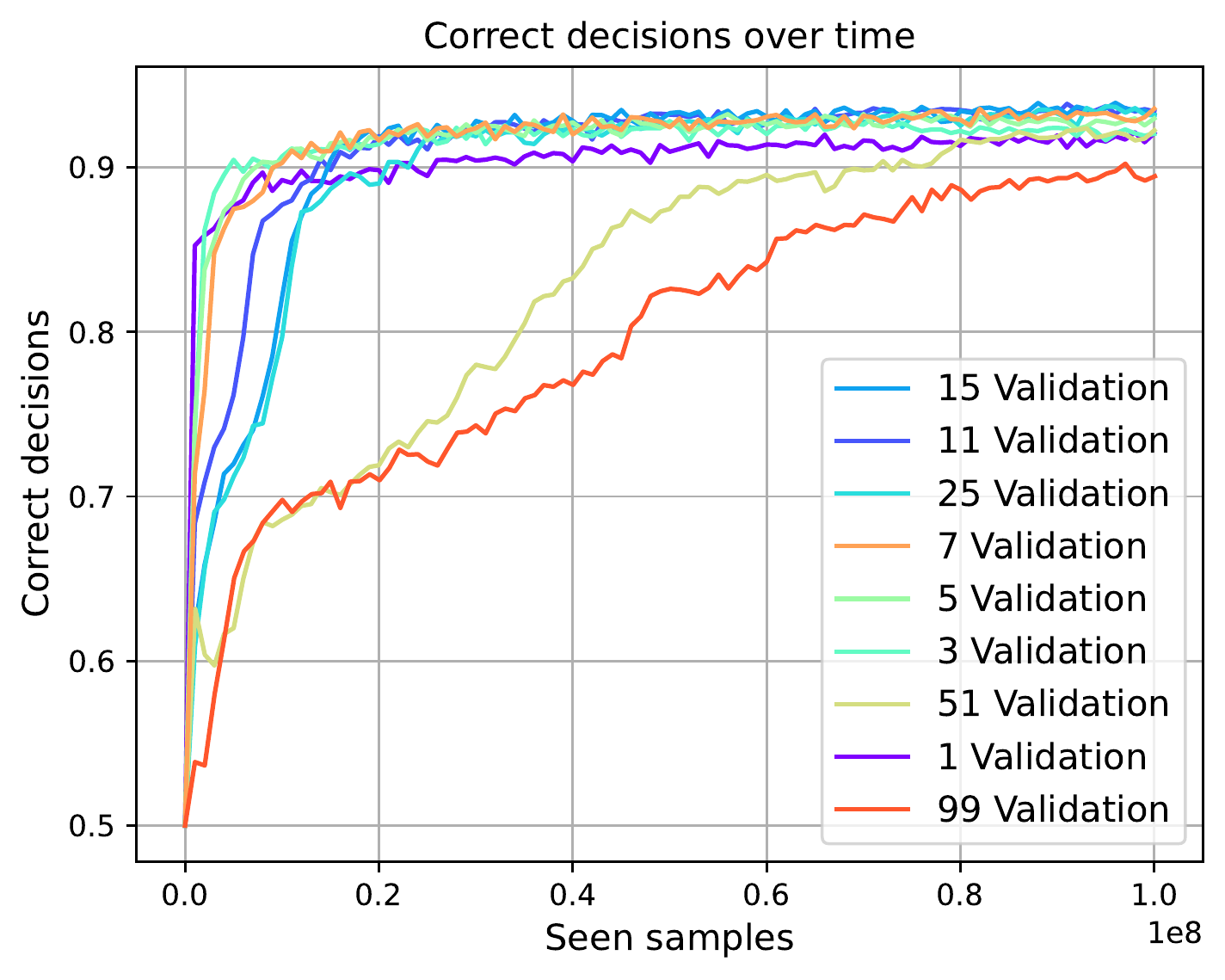}
    \caption{Correct comparisons between two poker hands on the test dataset over time. Labels are sorted by highest peak-performance achieved.}
    \label{fig:poker_comparison}
\end{figure}

Figure \ref{fig:poker_comparison} shows the test accuracy of networks trained with different numbers of validations over the course of training. In this experiment, labels are sorted by their peak performance reached. 
We can observe that smaller validation numbers increase the performance of the learning faster in the early training but generally achieve lower peak performances. 
It seems that an appropriate choice of validation numbers crucially decides the performance of the learning: while too much re-sampling leads to greatly ineffective training, training without re-sampling and relying on a larger sample size also significantly hurts the peak performance.


Following this motivation to study the problem, we now formalize the problem we are researching here.

\section{Problem definition and terminology}
\label{section: Problem definition}

The problem we want to handle can be defined as follows:
The learner is trained with examples $\mathcal{X} = \{\mathbf{x}_i\}, i = 1\dots n$, which are associated with labels $\mathcal{L} = \{\lambda_j\}, j = 1\dots l$ according to an unknown function $f:\mathcal{X} \rightarrow \mathcal{L}$, with $y_{i} = f(\mathbf{x}_i)$.
We further assume that the labeling process is non-deterministically guided by a noisy oracle $\oracle$ with a probability vector $\mathbf{p} = (p_1, \dots, p_l)$, which assigns the example the correct label $\lambda_j$ with a probability $p_j$, and one of the other labels with a probability of $p_k$, where $p_j > p_k$ for all $k \neq j$. Thus, $y_i = \lambda_j$ with $j = \arg \max_k p_k$.

We will often abbreviate this notation by using $q = p_j$ for the probability of picking the correct label, and $w = 1-q = \sum_{k\neq j} p_k$ for picking one of the incorrect labels.

In this setting, we are given a budget $\budget$ of queries to the oracle $\oracle$, which we can use for creating a training set $\mathcal{T}$. As explained above, each query to $\oracle$ returns the correct label for a given example with probability $q$, and an incorrect label with probability $w$.
However, we can query the oracle $v$ times for the same sample, each time receiving a newly generated label which may or may not be correct according to $\mathbf{p}$. Such repeated queries for the same example are called \emph{validations},
as we are trying to validate whether the label is correct. Each validation counts as a new
query to the 
budget $\budget$ in the same way as queries for new examples. This
means that we can use the budget $\budget$ to create a training set consisting of, e.g., $|\mathcal{T}| = \budget$ different samples, each validated only once ($v = 1$), or we can compute $|\mathcal{T}| = \frac{\budget}{3}$ samples where each label has been validated three times. In general, the number of validations can be changed at any time. This means that one may choose it prior to training, change it over the course of training, or even change it depending on the current sample or its already received labels.

Validating labels reduces the sizes of the training data that the classifier will be able to use throughout training but increases the likelihood of samples receiving the correct label. After $v$ validations for the same example, in which each label has been observed with frequencies $\mathbf{v} = (v_1,\dots,v_l)$, this example is assigned a label $\lambda_v$ with a majority vote, i.e.,
\begin{equation}
\lambda_v = \lambda_k, \textrm{ with } k = \arg \max_j v_j.
\end{equation}
Ignoring ties between labels, the probability of $\lambda_v$ being the correct label $\lambda_j$ after $v$ queries is equal to the probability of this label being drawn more frequently than any of the other labels. 
The probabilities of observing the labels with frequencies $\mathbf{v}$ after $v$ validations are multinominal distributed with the probability vector $\mathbf{p}$, i.e., 

\begin{equation}
P(v_1, \dots, v_l) = \textrm{Mult}(\mathbf{v},\mathbf{p}).
\end{equation}
Thus, a lower bound for the probability that the correct label $\lambda_j$ is selected is the probability that $v_j > v_k, \forall k \neq j$:
\begin{equation}
    P(\lambda_j = \lambda_v) = 
    \sum_{\mathbf{v}: \sum_i v_i = v} \textrm{Mult}(\mathbf{v},\mathbf{p})
    \prod_{k\neq j}{\mathbbm{1}({v_j > v_k})}
\end{equation}
where $\mathbbm{1}(.)$ is the indicator function. 

\begin{figure}
    \centering
    \includegraphics[width = \linewidth]{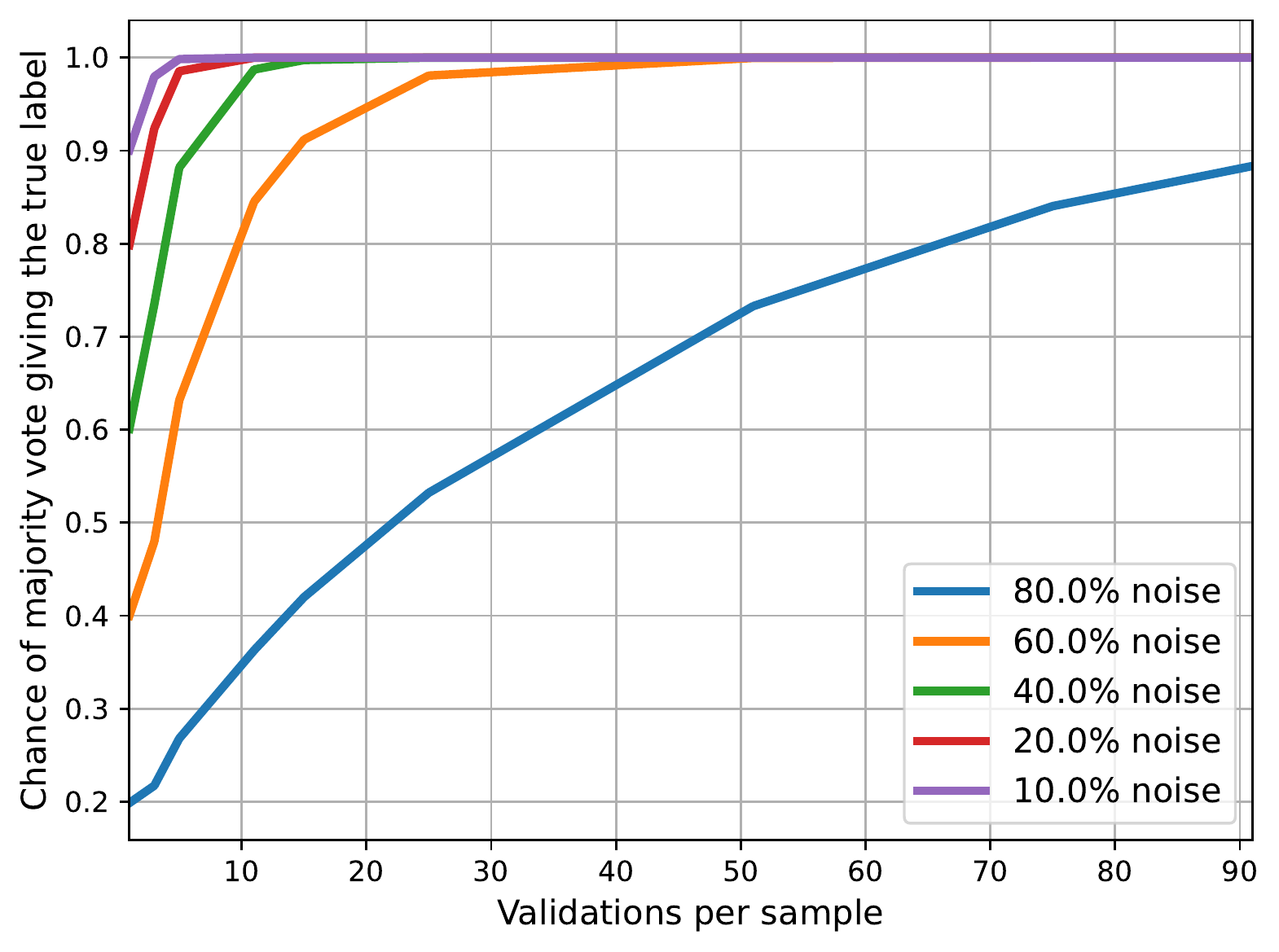}
    \caption{Probability of receiving the correct label given the number of validations for different probabilities of receiving incorrect labels. This simulation assumes 10 classes where all incorrect classes have equal chances.}
    \label{fig:mc_chance_label}
\end{figure}

Obviously, $P(\lambda_j = \lambda_v)$ depends on the number of validations $v$, the number of different labels $l$ and the chance of $\oracle$ returning the correct label $q$. For $\lim_{q \to 1}$, $\lim_{v \to \infty}$ and $\lim_{l \to \infty}$, the probability of receiving the correct label converges to 1 and the function has a logarithmic shape. Figure~\ref{fig:mc_chance_label} shows the probability of receiving a correct label from the oracle for varying numbers of validations for a dataset consisting of 10 different classes (e.g., like MNIST). Each curve corresponds to a noise level $w$. There, we also assume that all incorrect classes follow a uniform distribution, i.e.  $p_k = \frac{w}{l-1}$ for all $k \neq j$. The probabilities to receive the correct labels in domains with unequal noise differ.
From this, it is obvious that the choice of validations crucially depends on the noise in the data, as for lower noise few validations already result in almost perfect accuracy, while higher noise data benefits longer from additional validations to increase the reliability of the received labels.

\paragraph{Research Questions.}
On the basis of the above formalization,
we define a number of research questions concerning the label validation that should be answered in this work. Initially, it is not clear that any validation of labels is an efficient strategy, as re-evaluating the label of a known example is equally as costly as obtaining a label for a new example, which raises the question of whether such validations, which reduce the training set size but increase the label reliability, have any benefit at all. Intuitively, one might think that creating a sample only once will yield the highest efficiency of training, as the algorithm will use the maximum number of different examples for training and not ''waste'' any examples for validating samples.
However, we show in later experiments (Sections \ref{section:poolornot} and \ref{section:results_stable}) 
that this is not the case and that validating samples multiple times leads to faster and more efficient training of a neural network in our experiments. In cases with very high levels of noise, not validating labels even stopped the network from learning at all.

In summary, the research questions investigated in the remainder of the paper are the following:
\begin{itemize}
    \item [R1:] Can it be better to validate unreliable labels by pooling them with a majority vote or should all examples be used for training?
    \item[R2:] How should the number of validations be chosen in order to most efficiently train a neural network?
    \item[R3:] What is the relationship between the number of validations and the level of noise in the data?
    \item[R4:] Can an adaptive number of validations result in better efficiency than a stable number?
\end{itemize}

In the following, we will address these questions empirically in a study where we add artificial noise to the MNIST dataset so that we can systematically investigate the trade-off between re-sampling for increased label reliability and additional sampling for increased data volume.

\section{Experiments on MNIST}
\label{section:MNIST}

The domain in which we investigate the previously phrased questions is the popular image-dataset MNIST \cite{deng2012mnist}. We first describe the experimental setup and afterward the results that concern the research questions formulated above.

\subsection{Experimental Setup}

In this section, the setup for the conducted experiments is outlined.
The MNIST database consists of 60,000 gray-scale training images of size 28x28, for which we simulate a labeling process with a noisy oracle.  The 10,000 test images are not obstructed with noise and are used to compute the test accuracy of the networks.

\paragraph{Labeling.}

Each image falls into one of ten classes which represent the ten digits.
For this dataset, a noise level of $w$ is modeled by an oracle with returns the correct label with probability $q = 1-w$, while all other labels have a chance of $\nicefrac{w}{9}$ of being returned. Note that in this setting, due to the comparably small training set, multiple epochs of the whole training set are required. Therefore, in contrast to the online learning setting described in Section~\ref{section:Poker}, the network will see the same training examples multiple times. Due to the noisy setting, the network may also receive different labels for the same examples in each epoch. However, in order not to disturb the validation parameters, we do not save labels from previous epochs. Validations are independent each time a sample is generated, even though it may have been generated at a previous epoch.

\paragraph{Training procedure.}
To provide a baseline performance, as well as providing a structure for the experiments with noise, a neural network is constructed by optimizing its architecture on the basic MNIST-dataset without noise. In the end, the architecture of the network consists of 6 convolutional layers organized in 3 blocks. For the full architecture, refer to Appendix \ref{fig:network_architecture}. With this, a final testing accuracy of 99.4\% is achieved, which is deemed high enough compared to the current state-of-the-art performance of 99.9\% \cite{an2020ensemble}. The specific choices of hyperparameters, while of no great importance to the questions and results of this work, are outlined in Table \ref{table:Hyperparameters}. 

\begin{table}[ht]
\vspace{-4mm}
\caption{Chosen hyperparameters for the MNIST network}
\begin{center}
\resizebox{0.95\columnwidth}{!}{
\begin{tabular}{ll}
\hline
\rowcolor[HTML]{C0C0C0} 
\label{table:Hyperparameters}
\textbf{Hyperparameter}        & \textbf{Choice}    \\ \hline
Learning rate                  & 0.01               \\ 
Optimizer                      & AdamW              \\ 
Loss function                  & Cross Entropy Loss \\ 
Dropout                        & 0.25               \\ 
Batch normalization            & Yes                \\ 
Number of convolutional layers & 6                  \\ 
Activation function            & ELU               \\ 
Batch size                     & 256                \\ \bottomrule
\end{tabular}
}
\end{center}
\vspace{-3mm}
\end{table}

In all experiments with noisy MNIST, the main performance metric is the test accuracy of the network in relation to the number of seen training samples. Note that, as outlined in Section~\ref{section: Problem definition}, multiple validations of the same sample count the same as seeing a new sample. For example, if $v = 3$ validations are performed on each example, the algorithm needs $3\times256 = 768$ examples in order to produce a single batch update, whereas when no validations are performed ($v=1$), three batch updates can be performed from the same number of examples. In our plots, all versions will consume the same amount of examples (shown on the $x$-axis). It is important that this is a conservative assumption and overstates the computational effort of validation. In reality, validating a label is much faster than generating a new sample, passing it through the network, computing its loss, and back-propagating it. Therefore, the advantages of validation are even higher than the results show when the true wall-clock time required for training is used as a metric.

With the assumption that validating samples with a majority vote is the same as viewing a new sample, some considerations have to be made to allow for a fair comparison between different validation numbers. As all performances here are related to the number of seen samples, and as the batch size was held consisted in the experiments, a network trained with multiple validations will have made fewer gradient descent steps in the same "time" as one with one validation by a factor of $v$. Therefore, we multiply the base learning rate of 0.01 by the number of validations that are used for the respective training run. This again can be regarded as a disadvantage for the networks with validations, as a higher learning rate will make it harder to efficiently descend in the loss landscape. However, this adaptation was made to normalize the potential decrease in loss per number of seen samples.

\subsection{Utility of validation}
\label{section:poolornot}

We first investigate whether validations have any advantage over directly training on all of the sampled data. To this end, we compare two learners that see exactly the same data, which consists of examples each of which have been sampled $v$ times. The first version of the network uses all of the samples directly for training the network, whereas the second version first processes the samples in an internal validation process. Thus, the first network is trained on $N$ examples, whereas the second network is trained on $\nicefrac{N}{v}$ examples, which have been derived from the original $N$ examples via majority voting. In the plots, however, the performances of both versions are aligned on the $N$ examples that they both have received.

\begin{figure}
    \centering
    \includegraphics[width = \linewidth]{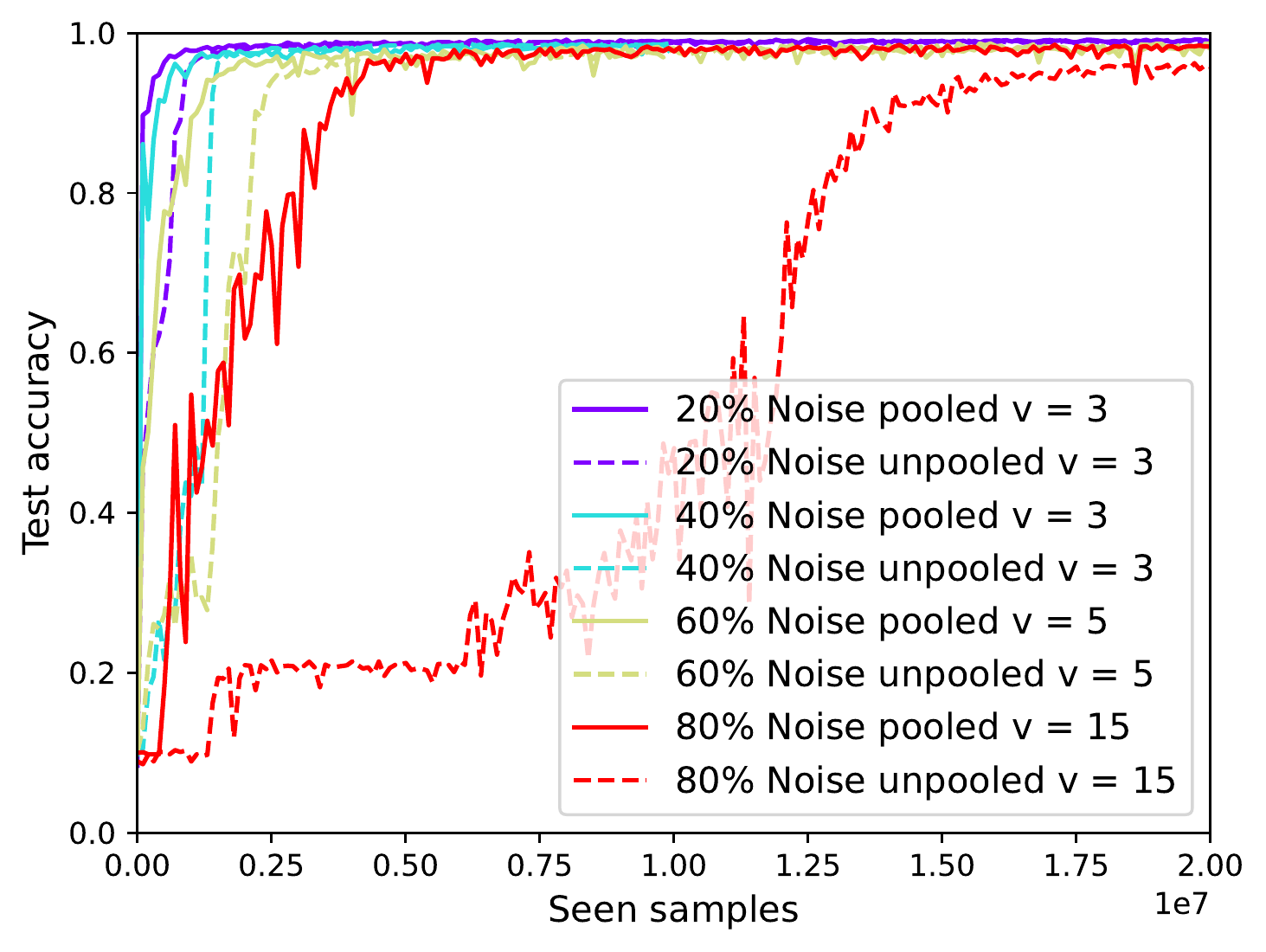}
    \caption{Comparison of processing samples with a majority vote (solid) and using all samples individually (dashed). In all cases, voting leads to faster training with better or similar peak performance.}
    \label{fig:pool_unpool}
\end{figure}

Figure \ref{fig:pool_unpool} shows this comparison of using all samples against first validating them with a majority vote for different noise values and different validation rates $v$. There, we see that in all settings, voting for the final label leads to better results than directly using all examples for training. This result makes intuitive sense, as the learner receives numerous incorrect labels in the former case and has to converge to a good classification performance by learning to ignore them. As such, it learns slower at the beginning of training. However, one may also make the argument that validating a single sample with a majority vote leads to a loss of information, as the classifier will only see the most often obtained label without any information about how often others were returned in queries. However, this experiment indicates that the former argument weighs stronger than the second one in practice.

One may argue that the learner would perform better if it could use $N$ different examples instead of repeated samples from validating. This is the setting that will be investigated in the next section, where we also try to investigate the relationship between validation amount and noise.

\subsection{Fixed validation choices}
\label{section:results_stable}

First, the performances of different constant choices for the validation parameter $v$ are compared for different noise levels in the training data. As learning curves for closely related performances can be hard to read, we use a different metric to evaluate how well a specific choice performed. Before training, several levels of test accuracy are chosen and performances are compared by the speed with which those levels were achieved. Levels ranged from very easy to reach (80\% test accuracy) to levels similar to the performance on non-noisy MNIST (99.3\% test accuracy). This metric of comparison also makes it possible to disregard overfitting as only the first time of reaching a level was important.

\begin{table}[ht]
\caption{Seen samples required to first reach the specified test accuracies (lower is better). Only the five best fixed values of $v$ are shown for conciseness. Full table in Appendix \ref{app:mnist_full}}
\label{tab:stable_val}
\begin{center}
\resizebox{\columnwidth}{!}{%
\begin{tabular}{lccccccc}
\hline
\rowcolor[HTML]{C0C0C0} 
 & 80\% & 90 \% & 95\% & 98\% & 99\% & 99.2\% & 99.3\%\\
 \hline
 \multicolumn{7}{c}{20\% Noise} \\
 \hline
 Fixed ($v=5$) & 250k & 550k & 950k & 1651k & 8150k & 20750k & 62950k \\
 Fixed ($v=3$) & 50k & 150k & 350k & 1050k & 5450k & 13150k & 63750k \\
 Fixed ($v=11$) & 652k & 1151k & 2051k & 3851k & 21051k & 28751k & NaN \\
 Fixed ($v = 7)$ & 451k & 851k & 1351k & 3250k & 12951k & 40951k & NaN \\
 Fixed ($v = 15)$ & 1053k & 1552k & 2651k & 5351k & 22352k & 48252k & NaN \\
  \hline
 \multicolumn{7}{c}{40\% Noise} \\
 \hline
 Fixed ($v=5$) & 350k & 650k & 950k & 2350k & 15350k & 28250k & 52450k \\
 Fixed ($v=11$) & 551k & 851k & 1450k & 3250k & 13452k & 26851k & 60850k \\
 Fixed ($v=7$) & 351k & 550k & 951k & 2350k & 12250k & 25250k & 74650k \\
 Fixed ($v=25$) & 2953k & 3853k & 4352k & 6553k & 29152k & 48752k & NaN\\
 Fixed ($v = 15)$ & 2953k & 3853k & 4352k & 6952k & 19552k & 60251k & NaN \\
   
  \hline
 \multicolumn{7}{c}{60\% Noise} \\
 \hline
 Fixed ($v=11$) & 352k & 851k & 1351k & 3151k & 11750k & 29451k & 45751k \\
 Fixed ($v=15$) & 951k & 1353k & 2153k & 4053k & 28153k & 40953k & NaN\\
 Fixed ($v=7$) & 1450k & 1750k & 2150k & 4050k & 22450k & NaN & NaN\\
 Fixed ($v=25$) & 1551k & 2754k & 4152k & 7551k & 27752k & NaN & NaN \\
 Fixed ($v=5$) & 1750k & 1850k & 2251k & 7550k & 45950k & NaN & NaN \\
  \hline
 \multicolumn{7}{c}{80\% Noise} \\
 \hline
 Fixed ($v=15$) & 1552k & 2552k & 3953k & 8353k & 42952k & NaN & NaN\\
 Fixed ($v=25$) & 2053k & 3553k & 4551k & 8754k & 45352k & NaN & NaN\\
 Fixed ($v=11$) & 1351k & 2351k & 3952k & 13651k & NaN & NaN & NaN\\
 Fixed ($v=51$) & 1958k & 3458k & 5553k & 13858k & NaN & NaN & NaN\\
 Fixed ($v=99$) & 3852k & 6852k & 11160k & 28752k & NaN & NaN & NaN\\
 \bottomrule
\end{tabular}%
}
\end{center}
\end{table}

Table \ref{tab:stable_val} shows an overview of this performance for four noise levels, 20\%, 40\%, 60\% and 80\%. Note that all test accuracies are computed on the full MNIST test dataset without any noise applied to it. The choices for the constant validation parameter were pre-defined as 1, 3, 5, 7, 11, 15, 25, 51, and 99 validations per sample. 

For this comparison, the five best-performing choices of $v$ are outlined. While some parameters performed obviously better than others, it is possible to interpret what makes one parameter better than another in several different ways, e.g. averaging over all obtained values, speed by which a specific level was reached, highest peak performance obtained, and others. We decided to order the performances by first using the speed by which the highest level, 99.3\% test accuracy, was reached. Then, all choices which did not reach this were ordered by the speed by which they achieved the next-lower level and continuing this process until a complete ordering is obtained. As such, we value peak performance the most, followed by the speed with which that performance was achieved. For the full overview of all tested parameters, refer to the Appendix \ref{app:mnist_full}.

In the 20\% noise setting, 5 and 3 validations performed best. Comparing the two, we see that $v=3$ was able to achieve the early goals more efficiently than $v=5$. However, the difference between both decreases throughout the different checkpoints until the 5-validation setting reaches the final performance slightly faster. We also see that the setting without validations, i.e.\ $v=1$, performed comparably poorly and was not able to reach the top 5. This confirms that investing some examples to obtain an increase in label quality at the expense of data volume provided a clear benefit here. In addition, we also see that the high-validation settings, i.e.\ 25, 51, and 99, were too inefficient for this environment. This observation can be explained by Figure~\ref{fig:mc_chance_label}, which shows how quickly label reliability increases with more validations in the 20\% noise setting. Therefore, querying the oracle for too many additional validations in a comparably low-noise environment wastes samples that should be used for new samples instead.

In the experiments with 40\% noise, using 5 validations per sample still performed best, but the order of efficiency for other choices differs. By increasing the noise of the oracle, $v=3$, the previous second-best choice, did not reach the top 5 and is replaced with 25. This observation underlines the hypothesis that with increasing noise of the oracle, the effectiveness of a higher number of validations for examples increases. We again see that in the earlier stages of training, lower validation numbers obtain vastly higher efficiencies. This advantage however is lost when training gets more difficult.

This trend continues when regarding the oracles with 60\% and 80\% noise. With 60\%, the previous top performer 5 falls drastically in performance. This is a result of the fact that the highest levels, 99.2\%, and 99.3\% test accuracy could not be achieved with lower than 11 validations per sample. In the highest noise setting, all of the five highest choices for $v$ performed best, again strengthening the previous observation that appropriate validation choice increases as noise increases. In addition, we find that in this extremely noisy environment, the low validation choices 1 and 3  were not able to learn at all and had the same test accuracy as classifying randomly. This shows that choosing an appropriate number of validation queries can make the difference between reaching test accuracies higher than 99\% and completely random classification of the learner. 

\clearpage
\paragraph{Observations.}
This series of experiments provides intriguing insights into the influence of the validation choice on the classifying performance. We observe that there seems to be an approximate range of appropriate parameter choice which differs depending on the noisiness of the oracle. In addition, we observe that these good choices seem to follow a normal distribution with decreasing performance when choosing too few or too many validations.

Secondly, we observe that too low validations do not reach the same peak performances but generally reach earlier performance levels more efficiently. Thus, we speculate that beginning training with lower validations and increasing them as classifying accuracy increases may be beneficial. We investigate this further in Section \ref{section:scheduled}.

Lastly, we want to emphasize the correlation between noise and validations. In contrast to the artificial environment studied here, we generally can not assume that the noisiness of the data is known in advance. Thus, approaches that are able to approximate noise and adapt validations based on that may perform better than fixed choices.

This is also the key difference between the experiments on Poker and MNIST. With MNIST, an even noise was applied to each training sample. In the Poker domain, noise naturally varies between training samples. While some hands have a very high chance of winning upward of 90\%, which results in many validations showing their win, other hands have equal winning chances. As such, the effectiveness of validations is different for different hands, which may make an appropriate choice much harder, again resulting in a need for adaptive solutions.

\subsection{Scheduled validation policies}
\label{section:scheduled}

Based on the results shown in~Table \ref{tab:stable_val}, we hypothesized that the importance of label reliability increases with improving network performance and training duration. In the early stages of training, the classifiers trained on the nosier samples increase in performance faster but are not able to reach higher performances in the later stages. Therefore, we speculate that starting training with a higher amount of lower-validated labels and switching to fewer but more reliable samples later can outperform the previously shown fixed approaches.

To provide a first investigation of this, we define several experiments in which the choice of label validation was switched at pre-defined points in training. These points do not depend on the returned label variance or network accuracy, as both of those measures can be highly unreliable, but we rather use fixed numbers of samples after which the choice of validations is changed to obtain more reliable labels. We call such variants, which have pre-defined switching points, "scheduled" policies.

For such scheduled approaches, we defined a starting number of validations to perform per sample, a range over which the validation number increases over time, and a maximum number of validations to perform. After the first 10\% of training on the initial parameter, the policy switches to the next-higher choice. There, again 10\% of training time is spent and the process repeats until it reaches the pre-defined maximum number where the remaining training time is used. As such, a classifier that is labeled "v = (1,3,5,7)" or short "v=($1 \dots 7$)" will have used labels that were validated with one, three, and five queries for the first, second, and third 10\% of training respectively while using labels with seven validations for the remaining 70\% of training time. We compare such scheduled approaches, and the statistical approach outlined in the next section, with the previously shown fixed variants in Table \ref{tab:mnist_best}.

\begin{table}[ht]
\caption{Seen samples required to first reach the specified test accuracies (lower is better). Only the five best overall results are shown. The schedules progress in the specified subranges of $v=(1,3,5,7,11,15,25,51,99)$. Full table in Appendix \ref{app:mnist_full}}
\label{tab:mnist_best}
\begin{center}
\resizebox{\columnwidth}{!}{%
\begin{tabular}{lccccccc}
\hline
\rowcolor[HTML]{C0C0C0} 
 & 80\% & 90 \% & 95\% & 98\% & 99\% & 99.2\% & 99.3\%\\
 \hline
 \multicolumn{7}{c}{20\% Noise} \\
 \hline
 Chi ($p < 0.05$) & 150k & 250k & 450k & 950k & 3450k & 12550k & 14950k \\
 Scheduled $(1\dots 5)$ & 50k & 150k& 250k & 850k & 5850k& 17350k & 27250k \\
 Scheduled ($1 \dots 3$) & 50k & 150k& 250k& 850k& 5550k & 12250k& 41950k \\
 Fixed ($v=5$) & 250k & 550k & 950k & 1651k & 8150k & 20750k & 62950k \\
 Fixed ($v=3$) & 50k & 150k & 350k & 1050k & 5450k & 13150k & 63750k \\
  \hline
 \multicolumn{7}{c}{40\% Noise} \\
 \hline
 Chi ($p < 0.05$) & 150k & 250k & 450k& 1251k & 6250k & 18051k & 34950k \\
 Scheduled ($1\dots 11$) & 50k & 650k & 850k & 2550k& 20450k& 26450k & 38350k \\
 Scheduled ($1 \dots 15$) & 50k & 250k & 350k & 2350k& 20350k & 27850k & 40951k \\
 Fixed ($v=5$) & 350k & 650k & 950k & 2350k & 15350k & 28250k & 52450k \\
 Scheduled ($3 \dots 11$) & 50k & 450k & 650k & 2950k & 10650k &22350k & 52450k \\
 \hline
 \multicolumn{7}{c}{60\% Noise} \\
 \hline
 Fixed ($v=11$) & 352k & 851k & 1351k & 3151k & 11750k & 29451k & 45751k \\
 Scheduled ($1 \dots 25$) & 50k & 1650k & 2150k & 11250k & 20050k & 21551k & NaN\\
 Scheduled ($1 \dots 15$) & 50k& 650k & 1250k & 10350k & 20252k & 39750k & NaN\\
 Fixed ($v=15$) & 951k & 1353k & 2153k & 4053k & 28153k & 40953k & NaN\\
 Scheduled ($1 \dots 7$) & 50k & 1350k & 1550k & 10250k & 21951k & Nan & NaN\\
  \hline
 \multicolumn{7}{c}{80\% Noise} \\
 \hline
 Scheduled ($11 \dots 51$) & 1650k & 2351k & 4051k & 10051k & 22554k & NaN & NaN\\
 Scheduled ($7 \dots 25$) & 3250k & 8751k & 10051k & 12351k & 29052k & NaN & NaN\\
 Scheduled ($5 \dots 25$) & 7451k & 10050k & 10150k & 15052k & 29353k & NaN & NaN\\
 Scheduled ($3 \dots 25$) & 11250k & 13150k & 15251k & 19850k & 34654k & NaN & NaN\\
 Fixed ($v=15$) & 1552k & 2552k & 3953k & 8353k & 42952k & NaN & NaN\\
 \bottomrule
\end{tabular}%
}
\end{center}
\end{table}

These results seem to underline the hypothesis that starting with noisier samples and saving more validations for later improves the efficiency of training. In all noise categories, apart from a single outlier in the 60\% noise setting, using the scheduled switches improved the efficiency with which the accuracy levels were reached. Additionally, in almost all cases, the scheduled variant ending on $X$ validations performed better than the one with $X$ fixed validations. We also observe that in general, choosing higher parameters for the last validation value performed better than lower ones, but this may be an artifact due to the specific choices made. Choosing too extreme values would likely break this observation.

\subsection{Dynamic validation policies}

As the last variant, we investigate how we can use the already obtained validation results at a given point to decide whether additional validations are required. The goal here is to query for additional labels until enough confidence in the obtained samples is achieved, such that further validations are not required. As is assumed in Section \ref{section: Problem definition}, $\forall_{k\neq j}: p_j > p_k$. Therefore, one policy is to stop querying the oracle for labels if the obtained labels \textbf{v} give sufficient reason to believe that $\arg \max_i v_i = j$, which also implies that the distribution from which \textbf{v} was samples has the highest probability at $p_j$. 

As this problem is not trivial to solve, we approximate it with the problem of deciding whether \textbf{v's} distribution is sufficiently different from a uniform distribution, which can be achieved with the chi-square goodness of fit test. Thus, we pre-define a confidence threshold and then query the oracle for labels until the p-value of the chi-square goodness of fit test on the returned labels falls below the threshold, thereby modeling that we are sufficiently certain that the underlying distribution of query results is not uniform. See Algorithm \ref{pseudo:chi_square_validation} for a pseudo-code of this process.

\begin{algorithm}
\caption{Chi-square validation}\label{pseudo:chi_square_validation}
\begin{algorithmic}[1]
\STATE $s \gets \text{sample to validate}$
\STATE $c \gets \text{confidence threshold}$
\STATE $l \gets \text{number of classes}$
\STATE $counts \gets \text{array with zeros of length l}$
\STATE $p \gets 1$

\WHILE{$p > \text{confidence threshold}$}
    \STATE $\lambda \gets \oracle(s)$
    \STATE $\text{counts}[\lambda] \text{+=} 1$
    \STATE $p \gets \text{chi square(counts)}$
\ENDWHILE

\STATE $\textbf{return } \text{argmax(counts)}$

\end{algorithmic}
\end{algorithm}

This algorithm has two key disadvantages; sample sizes can be extremely small and the algorithm will conclude querying if the label distribution is assumed to be non-uniform, but does not have a clear peak. As appropriate validation numbers in all experiments ranged between 3 and 25, the former problem can not be solved with any statistical test, as they will generally not work reliably with such small sample sizes. However, we do not require very high accuracies of when to stop querying, and the network can learn while receiving numerous incorrect labels. The task of this algorithm is not to be as accurate as possible, but rather to provide an approximation of how many validations are required for the given setting. To illustrate the second problem, consider the following two counts of labels received by validating a sample of the MNIST-dataset ten times: [0,0,7,0,1,0,2,0,0,0] and [0,0,5,0,0,0,0,0,5,0].

Both of those provide very high confidences, i.e., low p-values of 0.000001411 and 0.000007599 respectively, in rejecting a uniform distribution. However, while we can reasonably predict label 2 for the first case, it is not clear which label should be used for the latter one. While making incorrect decisions in such cases will likely influence the performance of the learning, much more computationally intensive practices would be required to avoid this problem. Table \ref{tab:mnist_best} includes the performances of this approach with a confidence threshold of 0.05. In the settings with lower noise, this worked extremely well, which makes intuitive sense. In the 20\% noise setting, labels are very often correct. If the oracle returns two of the same labels at the start, it is already unlikely that it was not the true label of that sample ($9 \cdot (\frac{0.2}{9})^2 \approx 0.4\%$) and only two validations are required to identify the likely true label. Additional queries are therefore only needed if at least one incorrect label was retrieved among the first two. We find that, while not perfect, mean validations (2.99, 4.93, 10.59, and 58.30 for the four levels of noise) are reasonable validation choices for each setting. However, the chi-square method was only able to achieve the best performances in the two lower-noise experiments, which leads us to speculate that the spread of obtained validations per sample was too large in the other two, as standard deviations were much higher there (9.2 and 64.36 compared to 1.55 and 3.49). A deeper investigation of this problem and other potential solutions is required.
´
\section{Conclusion}
\label{sec:conclusions}
In this paper, we have investigated the trade-off between label reliability and sample size. Our results have shown that in noisy or probabilistic domains, it is beneficial to re-sample the labels of known examples in order to increase their reliability, rather than obtaining new examples. 
Conversely, we have also seen that this advantage is lost if too many replicas of the same examples are re-sampled and that the optimal validation rate increases with increasing noise levels. Moreover, we have demonstrated the effectiveness of a scheduled switching of the validation policy, which increases the number of validations over time. This variant combines the advantages of low validation rates, which converge much faster in the initial training phases, and high validation rates, which are most effective for the final tuning of the network. Finally, we have also obtained first encouraging results on a dynamic, sample-dependent validation policy, which worked especially well in low and medium noise settings, but more work in this area is required. A more robust method of computing the reliability of obtained validations could increase the performance of such a dynamic variant further and reduce currently seen problems when handling very high-noise settings.

\newpage


\bibliography{bib}

\begin{thebibliography}{14}
\providecommand{\natexlab}[1]{#1}
\providecommand{\url}[1]{\texttt{#1}}
\expandafter\ifx\csname urlstyle\endcsname\relax
  \providecommand{\doi}[1]{doi: #1}\else
  \providecommand{\doi}{doi: \begingroup \urlstyle{rm}\Url}\fi

\bibitem[An et~al.(2020)An, Lee, Park, Yang, and So]{an2020ensemble}
An, S., Lee, M.~J., Park, S., Yang, H., and So, J.
\newblock An ensemble of simple convolutional neural network models for {MNIST}
  digit recognition.
\newblock \emph{arXiv}, 2008.10400, 2020.

\bibitem[Bertram et~al.(2021)Bertram, F{\"{u}}rnkranz, and
  M{\"{u}}ller]{bertram2021comparison}
Bertram, T., F{\"{u}}rnkranz, J., and M{\"{u}}ller, M.
\newblock A comparison of contextual and non-contextual preference ranking for
  set addition problems.
\newblock In \emph{Proceedings of ICML SubsetML workshop}, 2021.
\newblock arXiv:2107.04438.

\bibitem[Brodley \& Friedl(1999)Brodley and Friedl]{NoiseFiltering}
Brodley, C.~E. and Friedl, M.~A.
\newblock Identifying mislabeled training data.
\newblock \emph{Journal of Artificial Intelligence Research}, 11:\penalty0
  131--167, 1999.
\newblock \doi{10.1613/jair.606}.

\bibitem[Chen et~al.(2019)Chen, Liao, Chen, and Zhang]{chen2019understanding}
Chen, P., Liao, B., Chen, G., and Zhang, S.
\newblock Understanding and utilizing deep neural networks trained with noisy
  labels.
\newblock In Chaudhuri, K. and Salakhutdinov, R. (eds.), \emph{Proceedings of
  the 36th International Conference on Machine Learning (ICML)}, pp.\
  1062--1070, Long Beach, California, {USA}, 2019. {PMLR}.

\bibitem[Chicco(2021)]{chicco2021siamese}
Chicco, D.
\newblock Siamese neural networks: An overview.
\newblock In Cartwright, H.~M. (ed.), \emph{Artificial Neural Networks}, pp.\
  73--94. Springer, 3rd edition, 2021.

\bibitem[Deng(2012)]{deng2012mnist}
Deng, L.
\newblock The {MNIST} database of handwritten digit images for machine learning
  research.
\newblock \emph{IEEE Signal Processing Magazine}, 29\penalty0 (6):\penalty0
  141--142, 2012.

\bibitem[Koch et~al.(2015)Koch, Zemel, and Salakhutdinov]{koch2015siamese}
Koch, G., Zemel, R., and Salakhutdinov, R.
\newblock Siamese neural networks for one-shot image recognition.
\newblock In \emph{Proceedings of the ICML Deep Learning Workshop}. Lille,
  2015.

\bibitem[Reed et~al.(2015)Reed, Lee, Anguelov, Szegedy, Erhan, and
  Rabinovich]{reed2014training}
Reed, S.~E., Lee, H., Anguelov, D., Szegedy, C., Erhan, D., and Rabinovich, A.
\newblock Training deep neural networks on noisy labels with bootstrapping.
\newblock In Bengio, Y. and LeCun, Y. (eds.), \emph{Workshop Track Proceedings
  of the 3rd International Conference on Learning Representations ({ICLR})},
  San Diego, CA, USA, 2015.
\newblock arXiv:1412.6596.

\bibitem[Settles(2012)]{ActiveLearning}
Settles, B.
\newblock \emph{Active Learning}.
\newblock Synthesis Lectures on Artificial Intelligence and Machine Learning.
  Morgan {\&} Claypool Publishers, 2012.
\newblock \doi{10.2200/S00429ED1V01Y201207AIM018}.

\bibitem[Song et~al.(2020)Song, Kim, Park, and Lee]{song2020learning}
Song, H., Kim, M., Park, D., and Lee, J.
\newblock Learning from noisy labels with deep neural networks: {A} survey.
\newblock \emph{arXiv}, 2007.08199, 2020.

\bibitem[Sukhbaatar \& Fergus(2015)Sukhbaatar and
  Fergus]{sukhbaatar2014learning}
Sukhbaatar, S. and Fergus, R.
\newblock Learning from noisy labels with deep neural networks.
\newblock In Bengio, Y. and LeCun, Y. (eds.), \emph{Workshop Track Proceedings
  of the 3rd International Conference on Learning Representations ({ICLR})},
  San Diego, CA, USA, 2015.
\newblock arXiv:1406.2080.

\bibitem[Tesauro(1989)]{tesauro1989connectionist}
Tesauro, G.
\newblock Connectionist learning of expert preferences by comparison training.
\newblock In Touretzky, D.~S. (ed.), \emph{Advances in Neural Information
  Processing Systems 1}, pp.\  99--106, Denver, Colorado, 1989. Morgan
  Kaufmann.

\bibitem[Wang et~al.(2017)Wang, Zhang, Li, Zhang, and Lin]{costEffective}
Wang, K., Zhang, D., Li, Y., Zhang, R., and Lin, L.
\newblock Cost-effective active learning for deep image classification.
\newblock \emph{{IEEE} Transactions on Circuits and Systems for Video
  Technology}, 27\penalty0 (12):\penalty0 2591--2600, 2017.

\bibitem[Zhang \& Sabuncu(2018)Zhang and Sabuncu]{zhang2018generalized}
Zhang, Z. and Sabuncu, M.~R.
\newblock Generalized cross entropy loss for training deep neural networks with
  noisy labels.
\newblock In Bengio, S., Wallach, H.~M., Larochelle, H., Grauman, K.,
  Cesa{-}Bianchi, N., and Garnett, R. (eds.), \emph{Advances in Neural
  Information Processing Systems 31}, pp.\  8792--8802, Montr{\'{e}}al, Canada,
  2018.

\end{thebibliography}
\bibliographystyle{icml2022}

\newpage
\appendix
\onecolumn
\section{Full MNIST results}
\label{app:mnist_full}

\begin{table}[h]
\caption{Seen samples required to first reach the specified test accuracies (lower is better).}

\begin{center}
\resizebox{\columnwidth}{!}{%
\begin{tabular}{lccccccc}
\hline
\rowcolor[HTML]{C0C0C0} 
 & 80\% & 90 \% & 95\% & 98\% & 99\% & 99.2\% & 99.3\%\\
 \hline
 \multicolumn{7}{c}{20\% Noise} \\
 \hline
 Chi ($p < 0.05$) & 150k & 250k & 450k & 950k & 3450k & 12550k & 14950k \\
 Scheduled ($v = (1,3,5)$) & 50k & 150k& 250k & 850k & 5850k& 17350k & 27250k \\
 Scheduled ($v = (1,3)$) & 50k & 150k& 250k& 850k& 5550k & 12250k& 41950k \\
 Fixed ($v=5$) & 250k & 550k & 950k & 1651k & 8150k & 20750k & 62950k \\
 Fixed ($v=3$) & 50k & 150k & 350k & 1050k & 5450k & 13150k & 63750k \\
 Scheduled ($v = (1,3,5,7,11)$) & 50k & 150k& 250k& 1250K & 5850k& 24351k &NaN \\
 Fixed ($v=11$) & 652k & 1151k & 2051k & 3851k & 21051k & 28751k & NaN \\
 Fixed ($v=7)$ & 451k & 851k & 1351k & 3250k & 12951k & 40951k & NaN \\
 Fixed ($v=15)$ & 1053k & 1552k & 2651k & 5351k & 22352k & 48252k & NaN \\
 Fixed ($v=25)$ & 1450k & 2453k & 3652k & 7353k & 27953k & 71053k & NaN \\ 
 Fixed ($v=1)$ & 50k & 250k & 350k & 1050k & 28350k & NaN & NaN \\
 Fixed ($v=51)$ & 2258k & 3360k & 4953k & 10759k & 41855k & NaN & NaN\\
 Fixed ($v=99)$ & 2255k & 5170k & 8751k & 20759k & 98765k & NaN & NaN\\ 
  \hline
 \multicolumn{7}{c}{40\% Noise} \\
 \hline
 Chi ($p < 0.05$) & 150k & 250k & 450k& 1251k & 6250k & 18051k & 34950k \\
 Scheduled ($v = (1,3,5,7,11)$) & 50k & 650k & 850k & 2550k& 20450k& 26450k & 38350k \\
 Scheduled ($v = (1,3,5,7,11,15)$) & 50k & 250k & 350k & 2350k& 20350k & 27850k & 40951k \\
 Fixed ($v=5$) & 350k & 650k & 950k & 2350k & 15350k & 28250k & 52450k \\
 Scheduled ($v = (3,5,7,11)$) & 50k & 450k & 650k & 2950k & 10650k &22350k & 52450k \\
 Fixed ($v=11$) & 551k & 851k & 1450k & 3250k & 13452k & 26851k & 60850k \\
 Fixed ($v=7$) & 351k & 550k & 951k & 2350k & 12250k & 25250k & 74650k \\
 Fixed ($v=25$) & 2953k & 3853k & 4352k & 6553k & 29152k & 48752k & NaN\\
 Fixed ($v = 15)$ & 2953k & 3853k & 4352k & 6952k & 19552k & 60251k & NaN \\
 Fixed ($v = 3)$ & 50k & 350k & 650k & 2650k & 3055k & NaN & NaN\\
 Fixed ($v = 51)$ & 1958k & 3458k & 4951k & 13454k & 48556k & NaN & NaN\\
 Fixed ($v = 1)$  & 50k & 250k& 350k& 2450k& NaN & NaN &NaN\\
 Fixed ($v = 99)$ & 3472k& 8170k& 13273k& 29852k& NaN & NaN &NaN\\
   
  \hline
\end{tabular}%
}
\end{center}
\end{table} 

\begin{table}[h]
\begin{center}
\resizebox{\columnwidth}{!}{%
\begin{tabular}{lccccccc}
\hline
\rowcolor[HTML]{C0C0C0} 
 & 80\% & 90 \% & 95\% & 98\% & 99\% & 99.2\% & 99.3\%\\
 \hline
 \multicolumn{7}{c}{60\% Noise} \\
 \hline
 Fixed ($v=11$) & 352k & 851k & 1351k & 3151k & 11750k & 29451k & 45751k \\
 Scheduled ($v = (1,3,5,7,11,15,25)$) & 50k & 1650k & 2150k & 11250k & 20050k & 21551k & NaN\\
 Scheduled ($v = (1,3,5,7,11,15)$) & 50k& 650k & 1250k & 10350k & 20252k & 39750k & NaN\\
 Fixed ($v=15$) & 951k & 1353k & 2153k & 4053k & 28153k & 40953k & NaN\\
 Scheduled ($v = (1,3,5,7)$) & 50k & 1350k & 1550k & 10250k & 21951k & Nan & NaN\\
 Fixed ($v=7$) & 1450k & 1750k & 2150k & 4050k & 22450k & NaN & NaN\\
 Fixed ($v=25$) & 1551k & 2754k & 4152k & 7551k & 27752k & NaN & NaN \\
 Chi ($p < 0.05$) & 1751k & 1951k & 2451k & 4551k & 37050k & NaN & NaN\\
 Fixed ($v=5$) & 1750k & 1850k & 2251k & 7550k & 45950k & NaN & NaN \\
 Scheduled ($v = (1,3,5)$) & 50k & 1950k & 2650k & 10350k & NaN & NaN & NaN\\
 Fixed ($v=51$) & 2754k & 4254K & 7157k & 13454k & NaN & NaN& NaN \\
 Fixed ($v=99$) & 41564 & 8170k & 14160k & 30054k & NaN & NaN &NaN\\
 Fixed ($v=1$) & 50k & 850k& 1850k & NaN & NaN& NaN &NaN \\
 Fixed ($v=3$) & 1150k & 1450k & 3150k & NaN &NaN &NaN &NaN\\
  \hline
 \multicolumn{7}{c}{80\% Noise} \\
 \hline
 Scheduled ($v = (11,15,25,51)$) & 1650k & 2351k & 4051k & 10051k & 22554k & NaN & NaN\\
 Scheduled ($v = (7,11,15,25)$) & 3250k & 8751k & 10051k & 12351k & 29052k & NaN & NaN\\
 Scheduled ($v = (5,7,11,15,25)$) & 7451k & 10050k & 10150k & 15052k & 29353k & NaN & NaN\\
 Scheduled ($v = (3,5,7,11,15,25)$) & 11250k & 13150k & 15251k & 19850k & 34654k & NaN & NaN\\
 Fixed ($v=15$) & 1552k & 2552k & 3953k & 8353k & 42952k & NaN & NaN\\
 Fixed ($v=25$) & 2053k & 3553k & 4551k & 8754k & 45352k & NaN & NaN\\
 Fixed ($v=11$) & 1351k & 2351k & 3952k & 13651k & NaN & NaN & NaN\\
 Fixed ($v=51$) & 1958k & 3458k & 5553k & 13858k & NaN & NaN & NaN\\
 Fixed ($v=99$) & 3852k & 6852k & 11160k & 28752k & NaN & NaN & NaN\\
 Chi ($p < 0.05$) & 3655k & 7759k & 13056k & 43557k & NaN & NaN & NaN \\
 Fixed ($v=7$) & 3350k & 4450k & 28450k & NaN & NaN &NaN &NaN\\
 Fixed ($v=5$) & 7850k & 16750k & NaN & NaN & NaN &NaN &NaN\\
 Fixed ($v=1$) & NaN & NaN &NaN & NaN & NaN &NaN &NaN\\
 Fixed ($v=3$) & NaN & NaN &NaN & NaN & NaN &NaN &NaN\\
 
 \hline
\end{tabular}%
}
\end{center}
\end{table}
\clearpage

\section{MNIST level results as graphs}
\begin{figure*}[h!]
    \begin{subfigure}{\textwidth}
        \centering
        \includegraphics[width = \linewidth]{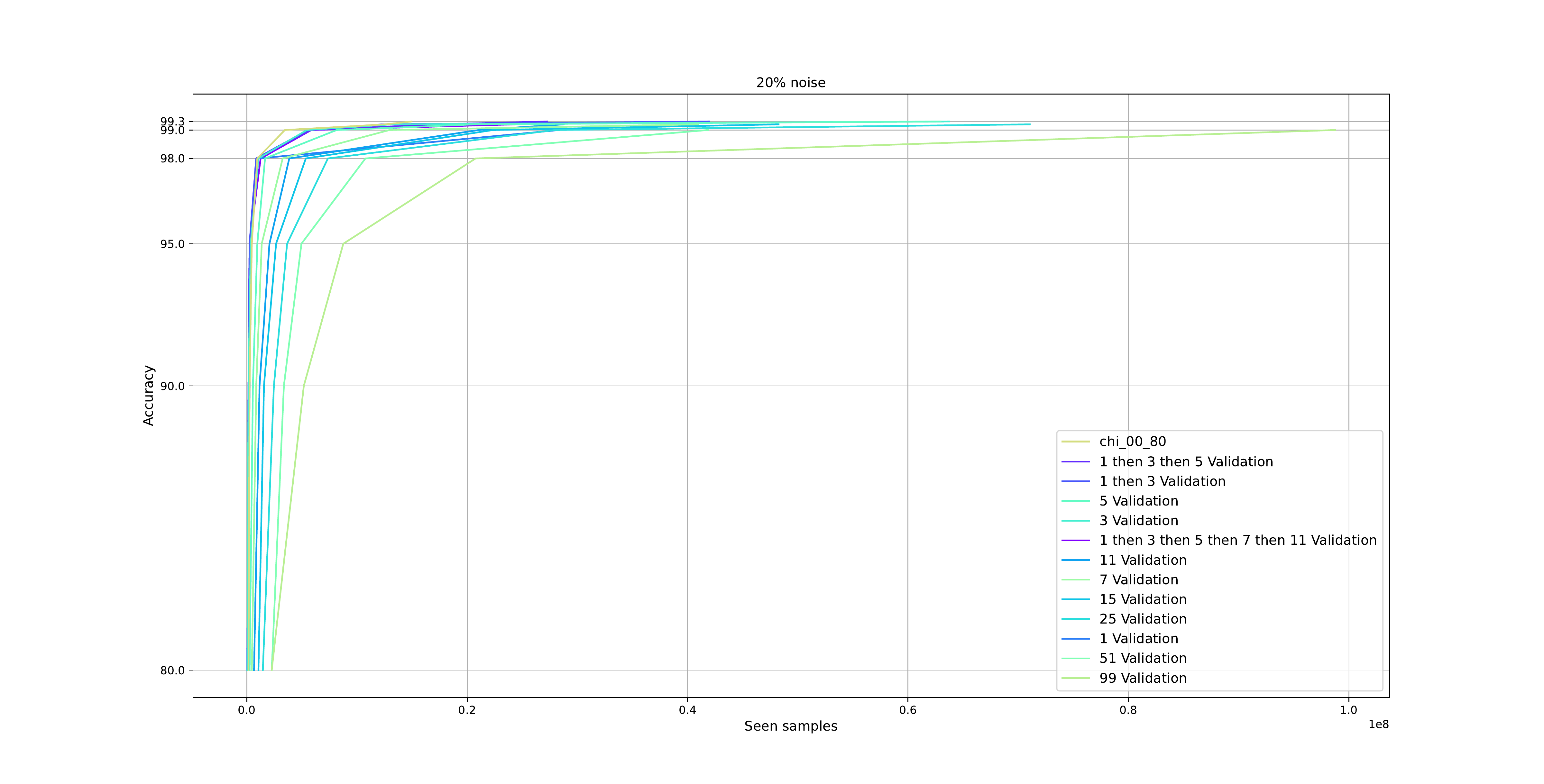}
            \caption{Accuracy checkpoints for 20\% noise}
        \label{fig:20_noise_stable_992}
    \end{subfigure}
    \begin{subfigure}{\textwidth}
      \centering
      \includegraphics[width = \linewidth]{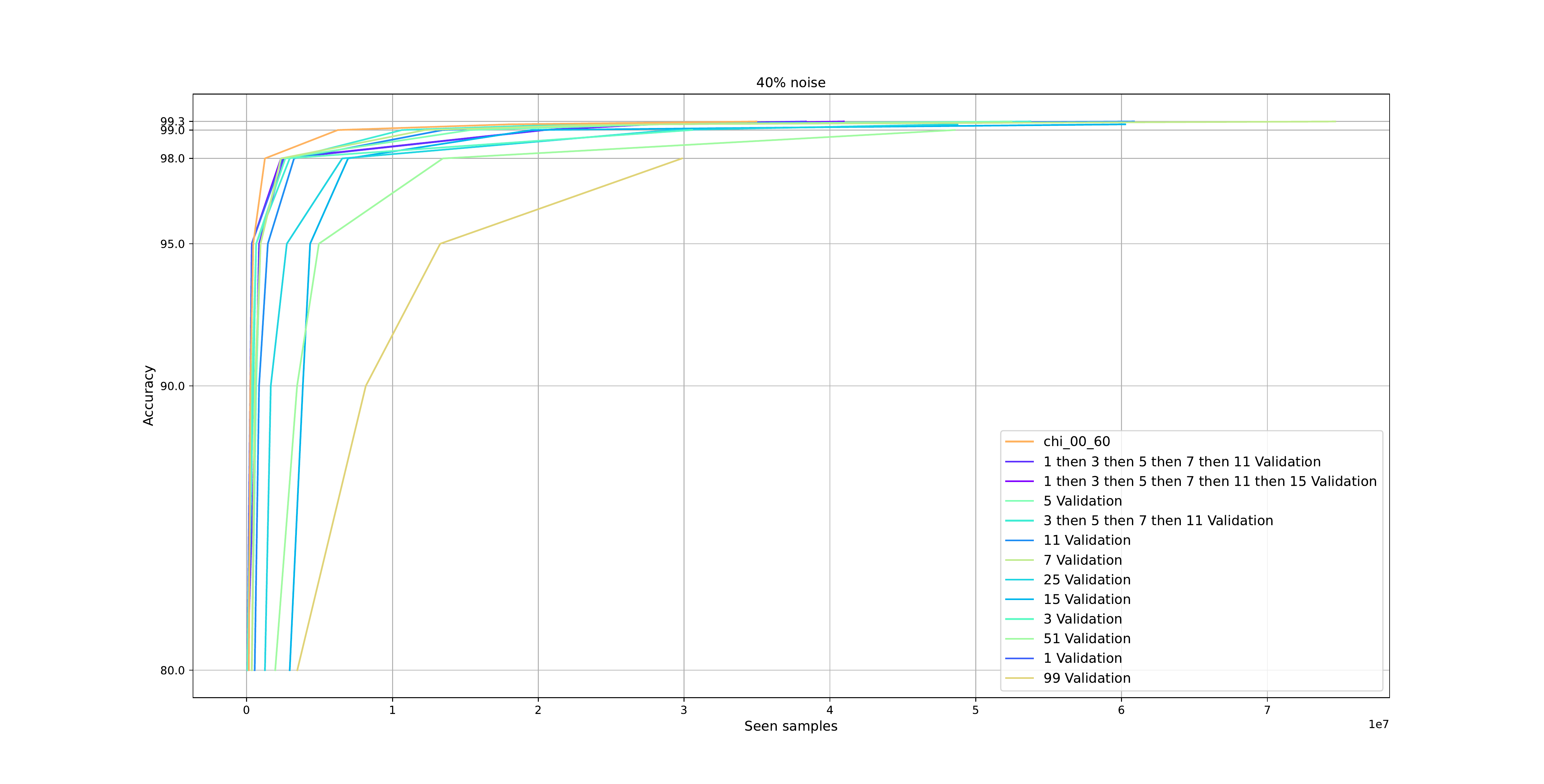}
            \caption{Accuracy checkpoints for 40\% noise}
      \label{fig:40_noise_stable_992}
    \end{subfigure}
\end{figure*}
\begin{figure*}\ContinuedFloat
    \begin{subfigure}{\textwidth}
        \centering
        \includegraphics[width = \linewidth]{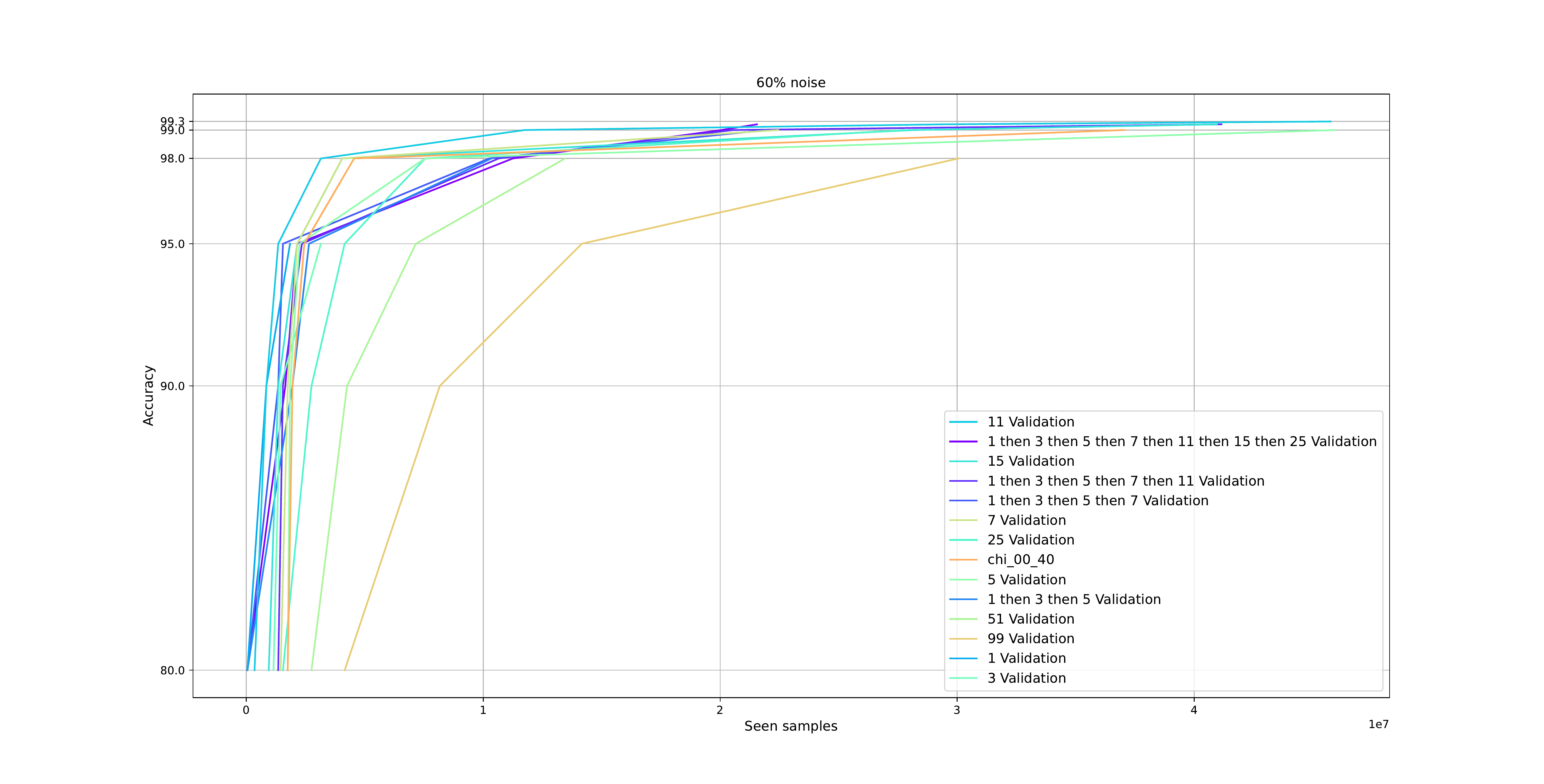}
            \caption{Accuracy checkpoints for 60\% noise}
        \label{fig:60_noise_stable_992}
    \end{subfigure}
    \begin{subfigure}{\textwidth}
      \centering
      \includegraphics[width = \linewidth]{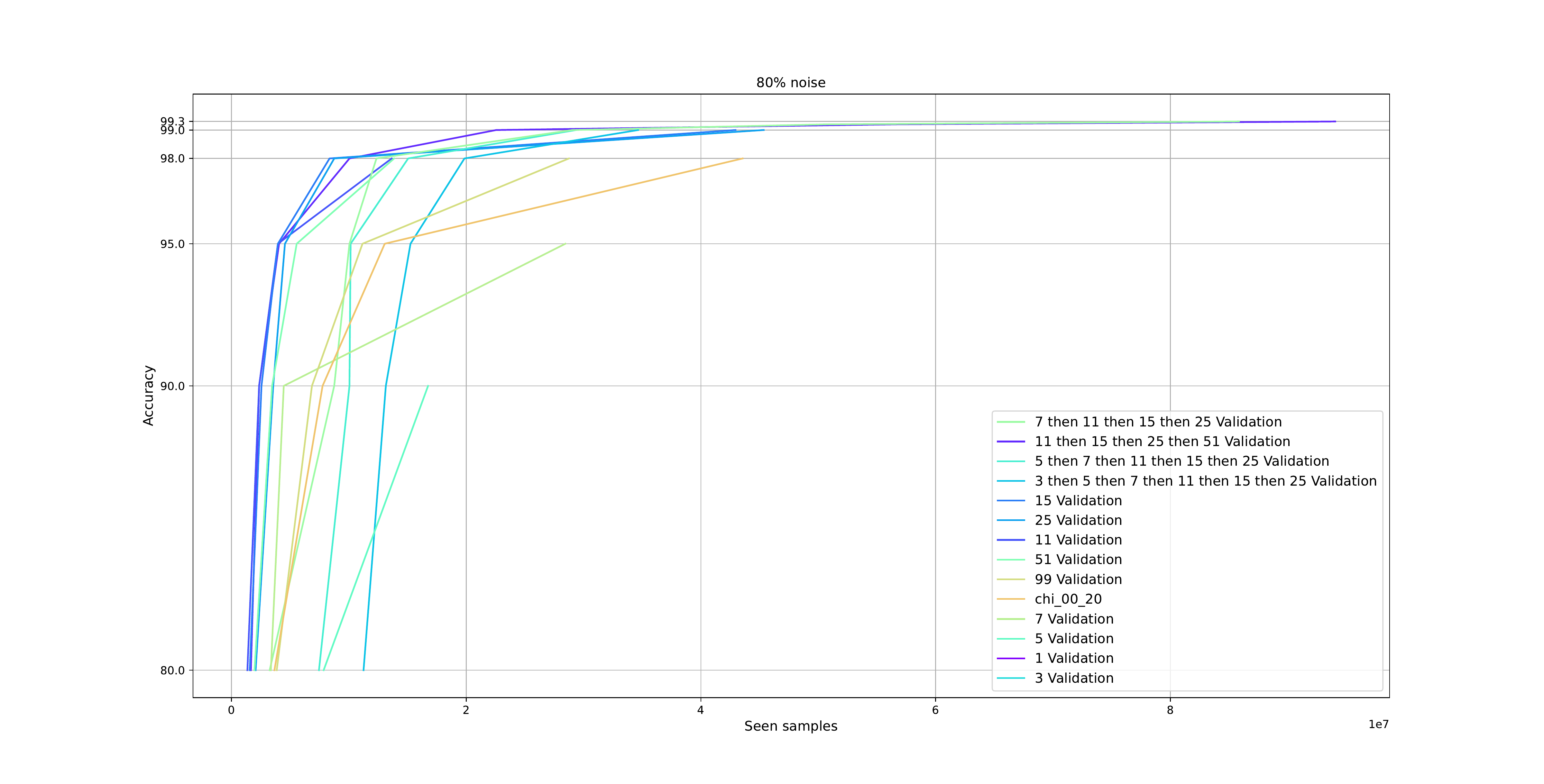}
            \caption{Accuracy checkpoints for 80\% noise}
      \label{fig:80_noise_stable_992}
    \end{subfigure}
    \caption{Comparing the different approaches by the speed with which they achieved pre-defined test accuracy checkpoint. This is a different representation of the data seen in Appendix \ref{app:mnist_full}}.
    \label{fig:comparison_noise_stable_992}
\end{figure*}

\clearpage
\section{MNIST training curves}
\begin{figure*}[h!]
    \begin{subfigure}{\textwidth}
        \centering
        \includegraphics[width = \linewidth]{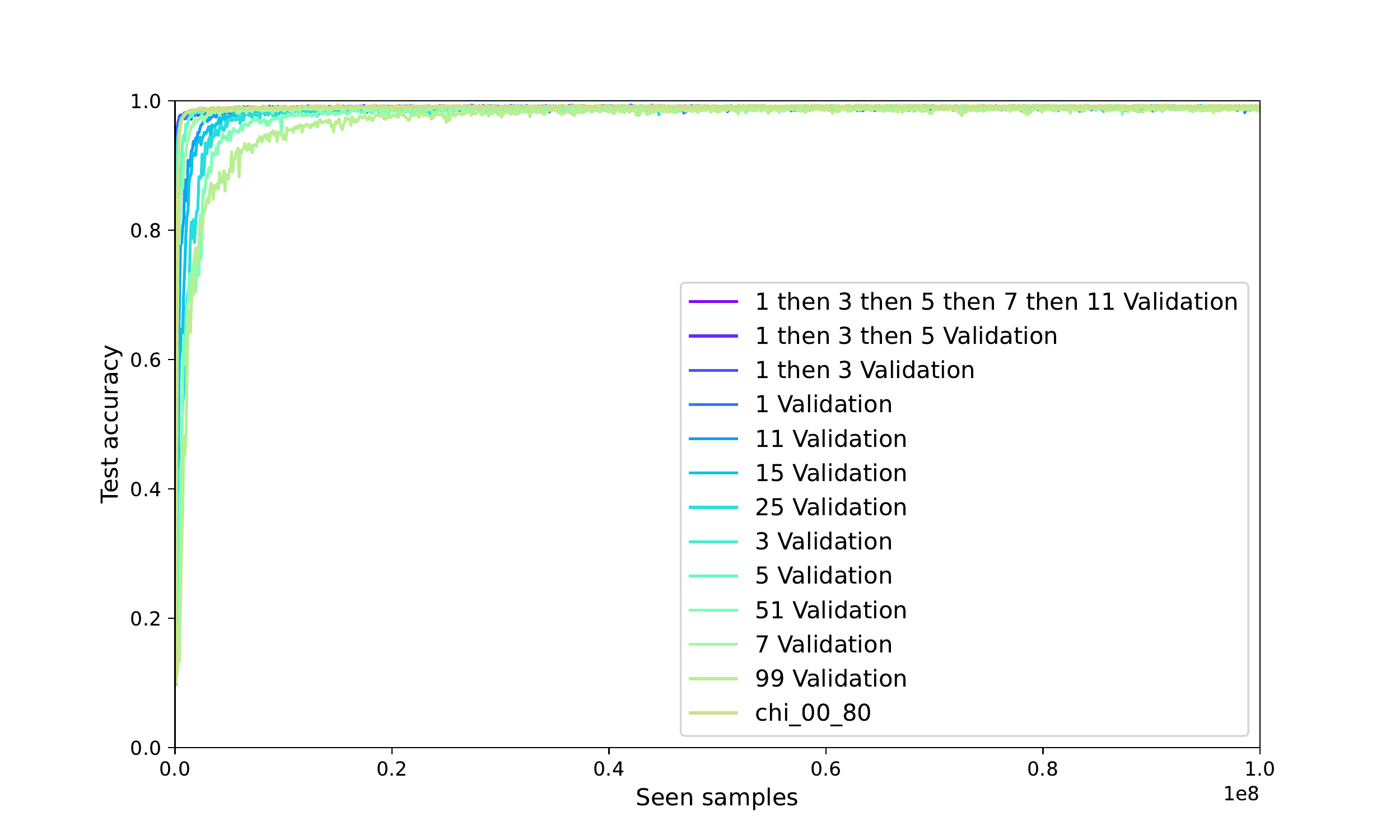}
            \caption{Training curves for 20\% noise}
    \end{subfigure}
    \begin{subfigure}{\textwidth}
      \centering
      \includegraphics[width = \linewidth]{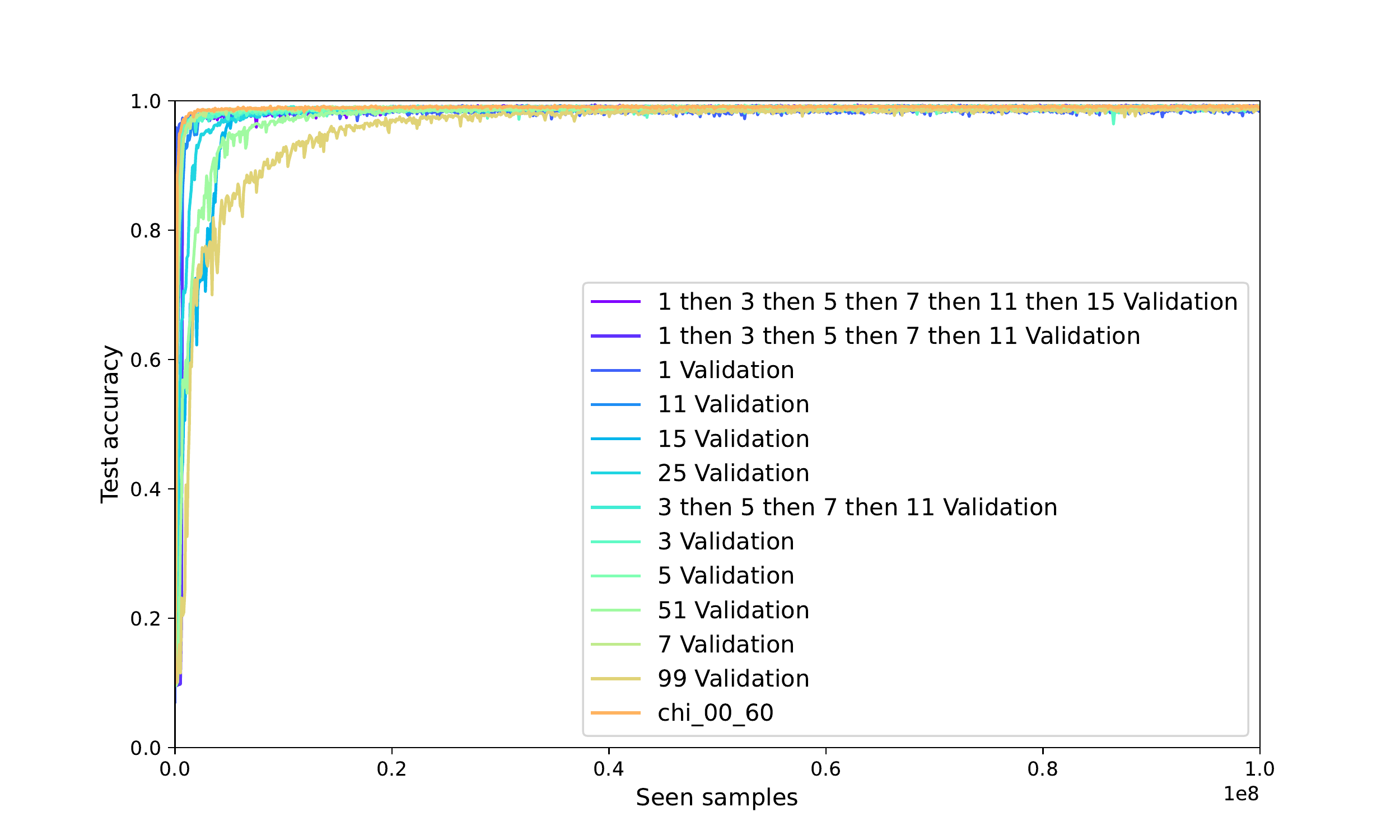}
            \caption{Training curves for 40\% noise}
    \end{subfigure}
    
\end{figure*}
\begin{figure*}
    \begin{subfigure}{\textwidth}
        \centering
        \includegraphics[width = \linewidth]{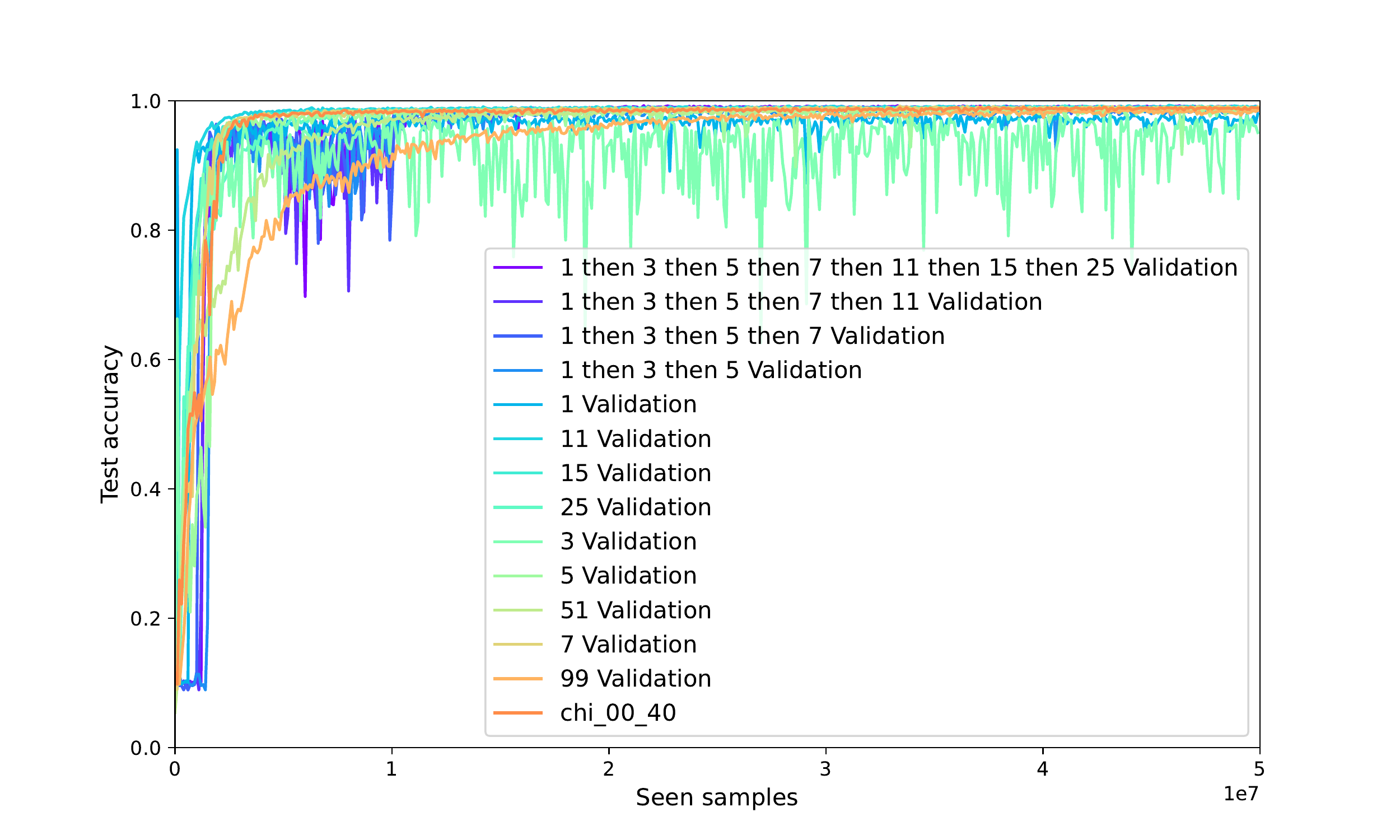}
            \caption{Training curves for 60\% noise}
    \end{subfigure}
    \begin{subfigure}{\textwidth}
      \centering
      \includegraphics[width = \linewidth]{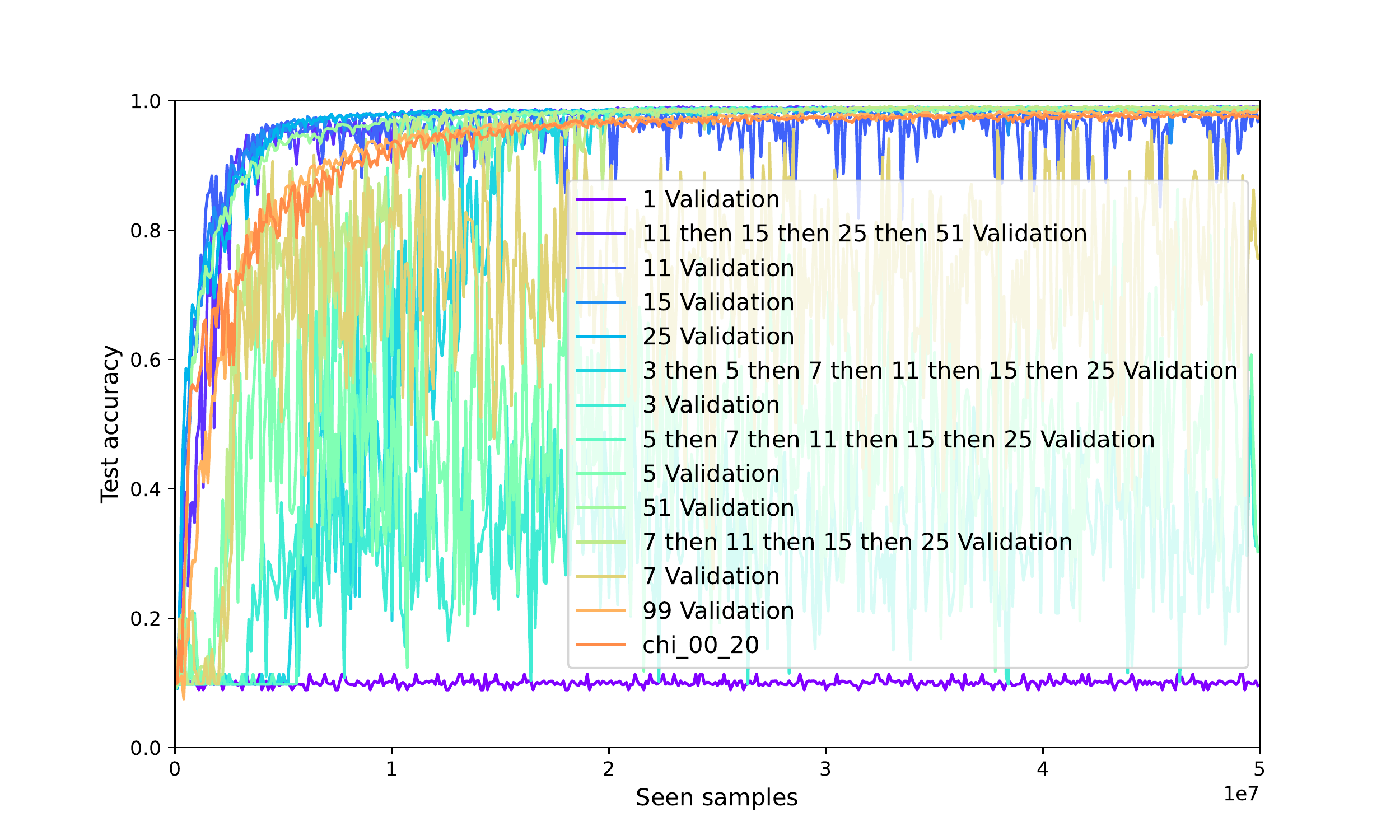}
            \caption{Training curves for 80\% noise}
    \end{subfigure}
    \caption{Test accuracy of trained networks over the course of training}
\end{figure*}

\section{MNIST network architecture}
\label{fig:network_architecture}
\begin{figure}[H]
    \centering
    \includegraphics[width = \linewidth]{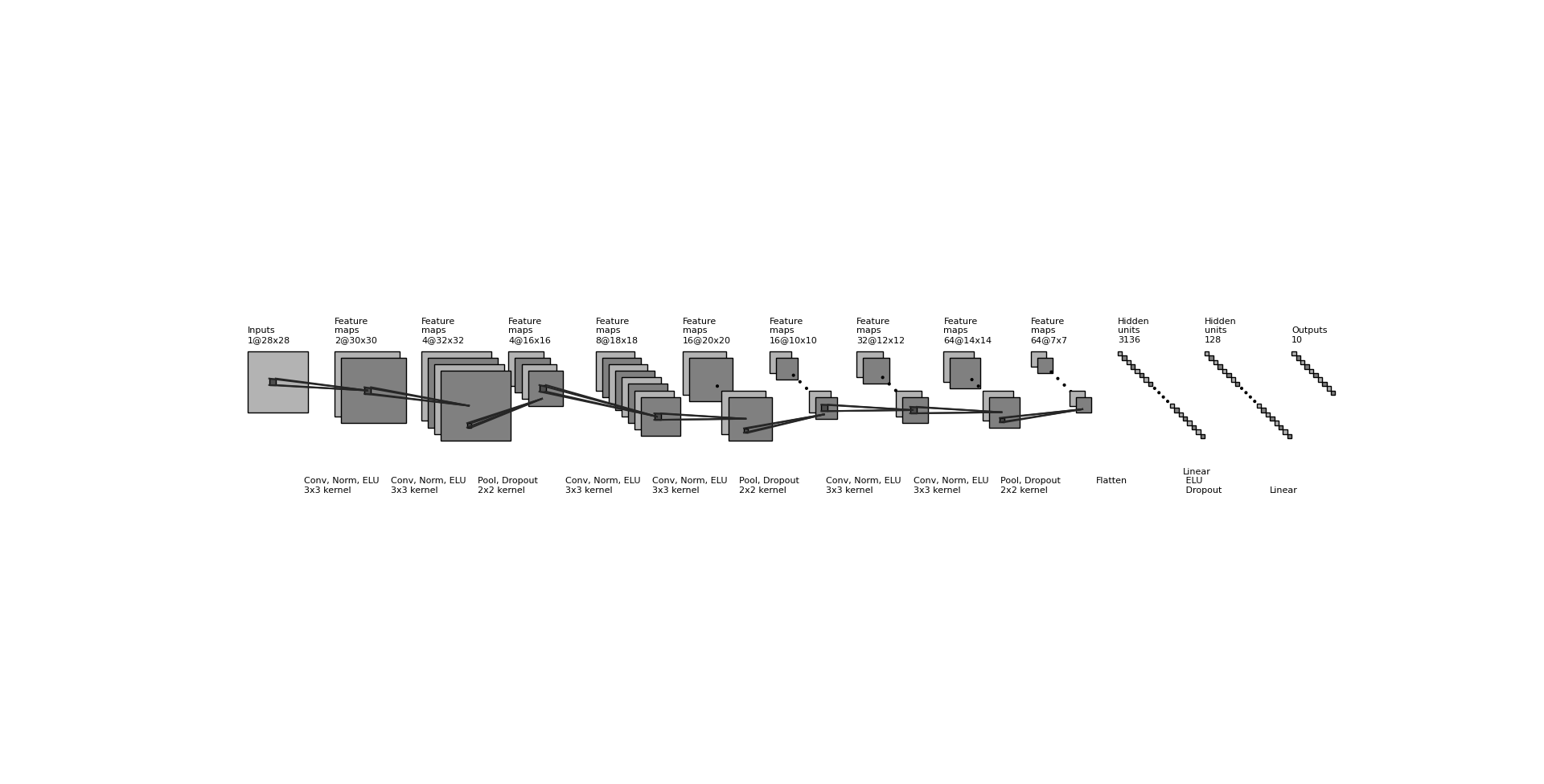}
\end{figure}

\end{document}